\DocumentMetadata{%
	pdfstandard=A-3u,
	pdfversion=1.7,
	lang=en-US,
}

\documentclass[balance,colorlinks,upint,subscriptcorrection,varvw,mathalfa=cal=boondoxo,spanish,german,vietnamese,russian,greek]{asmeconf}

\hypersetup{%
	pdfauthor={John H. Lienhard},									  
	pdftitle={ASME Conference Paper LaTeX Template},                  
	pdfkeywords={ASME conference paper, LaTeX template, BibTeX style},
	pdfsubject = {Describes the asmeconf LaTeX template},			  
	pdflicenseurl={https://ctan.org/pkg/asmeconf},
}

\begin{document}




\title{GenCAD-Self-Repairing: Feasibility Enhancement for 
3D CAD Generation} 
 
%
%
%

\SetAuthors{%
    Chikaha Tsuji\affil{1}\JointFirstAuthor, 
	Enrique Flores Medina\affil{1}\JointFirstAuthor,  
	Harshit Gupta\affil{1}\JointFirstAuthor\CorrespondingAuthor{harshit1@mit.edu},
    Md Ferdous Alam\affil{1}}

\SetAffiliation{1}{Department of Mechanical Engineering at Massachusetts Institute of Technology, Cambridge, MA}


\maketitle








\begin{abstract}

With the advancement of generative AI, research on its application to 3D model generation has gained traction, particularly in automating the creation of Computer-Aided Design (CAD) files from images. GenCAD~\cite{gencad} is a notable model in this domain, leveraging an autoregressive transformer-based architecture with a contrastive learning framework to generate CAD programs.

However, a major limitation of GenCAD is its inability to consistently produce feasible boundary representations (B-reps), with approximately 10\% of generated designs being infeasible. To address this, we propose \textbf{GenCAD-Self-Repairing}, a framework that enhances the feasibility of generative CAD models through diffusion guidance and a self-repairing pipeline. This framework integrates a guided diffusion denoising process in the latent space and a regression-based correction mechanism to refine infeasible CAD command sequences while preserving geometric accuracy. Our approach successfully converted two-thirds of infeasible designs in the baseline method into feasible ones, significantly improving the feasibility rate while simultaneously maintaining a reasonable level of geometric accuracy between the point clouds of ground truth models and generated models.

By significantly improving the feasibility rate of generating CAD models, our approach helps expand the availability of high-quality training data and enhances the applicability of AI-driven CAD generation in manufacturing, architecture, and product design.

\end{abstract}



\section{Introduction}
3D Modeling is a fundamental part of the complete product design process that allows for easy visualization of part shape, size, and configuration to be manufactured. In manufacturing, Computer-Aided Design (CAD) is particularly well-suited for 3D modeling due to its \textit{editability}, enabling precise geometric representation, parametric design modifications, and seamless integration with Computer-Aided Manufacturing (CAM) systems. However, the CAD modeling process is manual, intensive, and takes a lot of time and effort. To address this problem, and attenuate the time to design 3D models of products and parts, generative AI has been particularly useful~\cite{keskar2024driving}. 

Recent advancements in generative AI, on the other hand, have demonstrated remarkable capabilities in various domains, including text-to-image  \cite{Rombach_2022_CVPR} \cite{ramesh2022hierarchical}, text-to-3D \cite{poole2022dreamfusion} \cite{jun2023shap}, and even code generation \cite{chen2021evaluating} \cite{nijkamp2022codegen}. These models leverage large-scale datasets and deep learning architectures, particularly diffusion models~\cite{ho2020denoising} and transformers~\cite{vaswani2017attention}, to generate high-quality outputs. Given these successes, generative AI is increasingly being explored for automating engineering design tasks, including CAD generation.

However, despite its potential, the integration of generative AI with CAD modeling remains limited. Unlike meshes, voxels, and point clouds, which are commonly used in 3D generation tasks, CAD models require an editable and manufacturable representation that adheres to strict geometric and topological constraints. The industry-standard boundary representation (B-rep)~\cite{weiler1986topological} encodes parametric surfaces, edges, and vertices, making CAD models more structured but also more complex than other 3D formats. This hierarchical nature, combined with the need for precise topological relationships and parametric geometry, makes B-rep data challenging to process directly with standard neural network architectures. As a result, an intermediate, network-friendly representation is often required~\cite{gencad}, adding another layer of complexity to generative CAD modeling.

Another major challenge hindering research in generative CAD modeling is the lack of large-scale training datasets. Unlike image or text-based generative models, which benefit from vast publicly available datasets, CAD datasets are limited in size and diversity. Additionally, as highlighted in GenCAD~\cite{gencad} limitations, some existing CAD generation models do not always produce valid CAD outputs—some generated sequences fail to construct feasible designs, further restricting the availability of usable training data.

Therefore, this paper aims to improve the feasibility rate of CAD generation models while maintaining quality, thereby paving the way for expanding training datasets in future generative CAD modeling research.


\section{Related Work and Contribution} \label{sec:RelatedWork}
\begin{figure}[t!]
  \centering
  \includegraphics[scale=0.6]{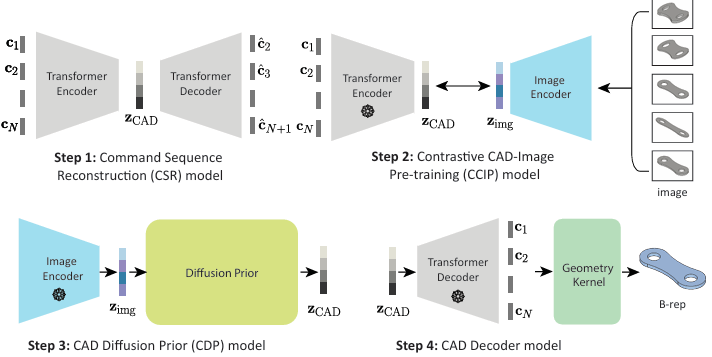}
      \caption{GenCAD~\cite{gencad} architecture (Reproduced from \cite{gencad}).}
  \label{fig:GencadArchitecture} 
\end{figure}

One major bottleneck in this field is the lack of large-scale, structured datasets for training AI models capable of generating realistic CAD designs. However, several datasets have been developed to support AI-driven CAD research. Among the few publicly accessible options, the ABC dataset is the largest, containing approximately one million CAD models sourced from various online repositories \cite{ABC}. This dataset contains B-reps but does not provide any design histories or labels. Later, the DeepCAD dataset was developed \cite{wu}. The DeepCAD dataset includes design histories by extracting details from CAD designs within the Onshape repository. Similarly, the Fusion 360 dataset offers design histories along with assembly information and face segmentation~\cite{willis}. However, its small size limits its effectiveness for training models that generalize well. The OpenECAD dataset uses a series of datasets that describe CAD operations via Python-like commands, where images include isometric, orthographic, and transparent views of 3D models~\cite{Yuan_2024}.

Building on these datasets, various AI-driven CAD generation models have been proposed. CAD Translator leverages both text and image encodings in an encoder-decoder framework to improve CAD reconstruction~\cite{li2024cad}. Other approaches, such as ComplexGen, view CAD reconstruction as a geometric primitive detection task, identifying vertices, edges, and surface patches to generate a valid B-rep structure \cite{R4}. Additionally, OpenECAD generates structured 2D sketches and 3D construction commands from images of 3D designs, and output 3D construction commands can pass through a 3D geometry kernel to create B-rep files \cite{Yuan_2024}.

\begin{figure}[t!]
\begin{center}
\includegraphics[scale=.45]{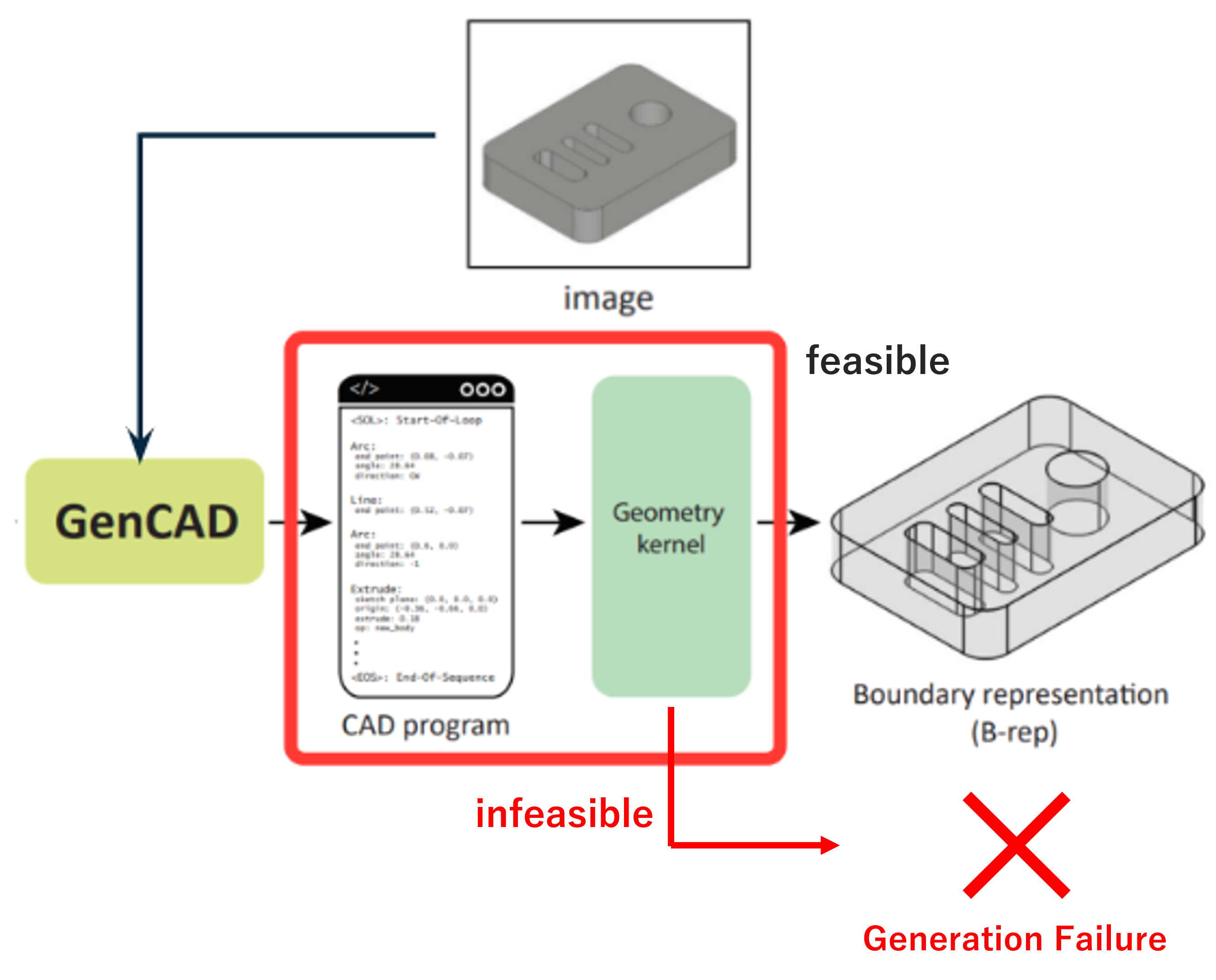}
\end{center}
\caption{GENCAD LIMITATION: Around 10\% of CAD programs generated from GenCAD~\cite{gencad} result in invalid geometry kernels (Modified from \cite{gencad}).}
\label{fig:GencadLimitation} 
\end{figure}

Among existing models, GenCAD~\cite{gencad} is the most closely related to our work. GenCAD~\cite{gencad} is a generative model that employs an autoregressive transformer-based architecture with a contrastive learning framework (Fig.~\ref{fig:GencadArchitecture}), enhancing the generation of CAD programs from input images and providing a representation learning framework for multiple data modalities relevant to engineering designs. GenCAD~\cite{gencad} is distinguished by learning the distribution of design histories, or command sequences, that make up a B-rep and has superiority over other state-of-the-art generative CAD design models~\cite{wu}~\cite{xu2022skexgen} in terms of generating command sequences, or a sequence of CAD commands and parameters design tokens, than B-rep files without design history. However, GenCAD~\cite{gencad} has a major limitation: it does not guarantee the successful conversion of all input images into valid CAD models (Fig.~\ref{fig:GencadLimitation}). Approximately 10$\%$ of the generated B-reps are infeasible, significantly limiting the usability of the model and the size of the GenCAD dataset.

To address this limitation, we propose GenCAD-Self-Repairing, a framework combining a guided diffusion denoising process in the latent space and a self-repair pipeline for infeasible CAD sequences. Our method not only improves the feasibility of AI-driven CAD generation, but also helps expand the existing training dataset for generative CAD modeling. Additionally, to validate the accuracy of the CAD models generated from the infeasible design space, divergence metrics such as Maximum Mean Discrepancy (MMD) have been calculated among the point clouds of the corrected CADs and the ground truth.

\begin{figure*}[t!]  
 \begin{center}

  \includegraphics[scale=0.5]{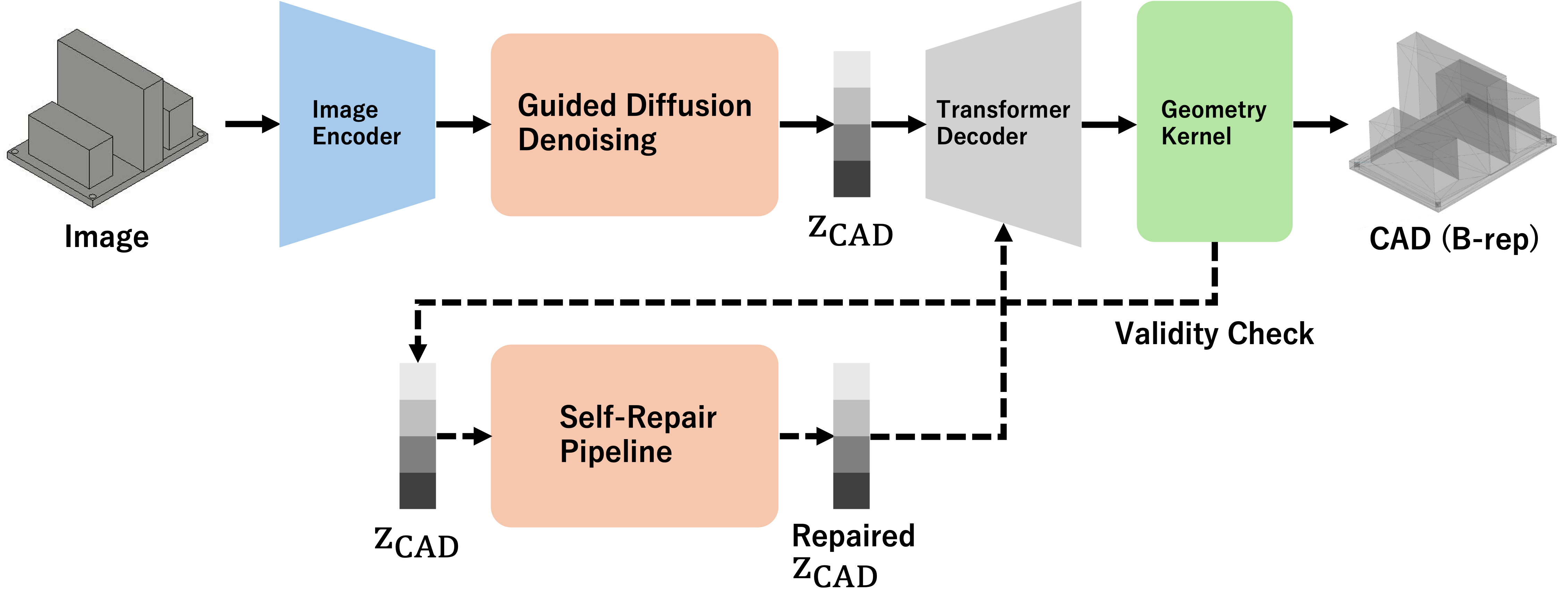}
  \vspace{-2mm}
  \caption{
Overview of GenCAD-Self-Repairing, which consists of a guided diffusion denoising process and a self-repair pipeline. The guided diffusion denoising directs latent vectors using classifier and regressor-based feedback, while the self-repair pipeline leverages an SSL Regressor to correct invalid latent representations. These mechanisms work together to improve the feasibility of CAD command generation.
}

  \label{fig:overview}
\vspace{-5mm}
  
 \end{center}
\end{figure*}

\section{methods} \label{sec:Method}

GenCAD-Self-Repairing consists of a guidance mechanism during the diffusion denoising process and a self-repair pipeline as shown in Fig.~\ref{fig:overview}.  The diffusion guidance incorporates feedback from a latent-space classifier (Latent CAD Classifier) and a Self-Supervised Learned Regressor (SSL Regressor), while the self-repair pipeline utilizes the same SSL Regressor used in the guidance. To train the Latent CAD Classifier, we first construct the GenCAD-Self-Repairing dataset.


The dataset creation process is discussed first, followed by the training of each machine learning model. Finally, we explain the rationale behind the proposed feasibility enhancement framework.


\begin{table}[t!]
\caption{
Summary of the GenCAD-Self-Repairing dataset. The dataset consists of images, latent representations labeled as valid or invalid, and ground truth latent representations derived from corresponding ground truth command sequences. 
}
\vspace{-3mm}
\begin{center}
\begin{tabular}{c c c}
\label{tab:dataset}
Type & Count & Shape  \\
\hline
Images & 133,617 & 1x448x448\\
Valid Latent with Labels & 123,809 & 1x257 \\
Invalid Latent with Labels & 9,808 & 1x257 \\
Ground Truth Latent & 26,723 & 1x256 \\
\hline
\end{tabular}
\vspace{-6mm}
\end{center}
\end{table}

\subsection*{3.1. GenCAD-Self-Repairing Dataset}
\begin{figure}[t!]
\begin{center}
\includegraphics[scale=.26]{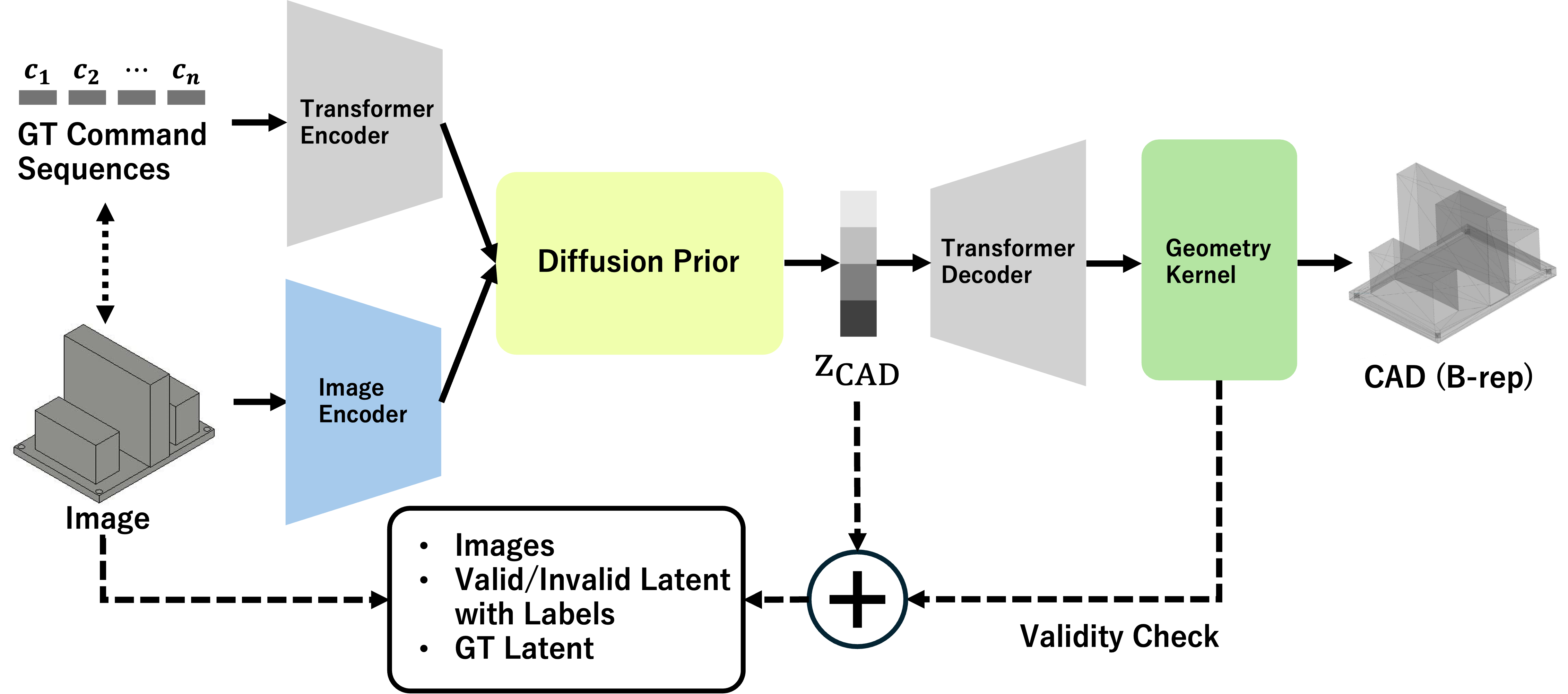}
\end{center}
\caption{Creation process OF the DATASET.}
\label{fig:dataset_classifier} 
\end{figure}

The GenCAD-Self-Repairing dataset stems from the GenCAD dataset \cite{gencad}. The summary of the dataset is shown in Table~\ref{tab:dataset} and the dataset creation process flow is illustrated in Fig.~\ref{fig:dataset_classifier}. 

It was created by randomly selecting 133,617 images from the total set of 840,947 - given the time and computing constraints - and then using the GenCAD model~\cite{gencad} to generate corresponding command sequences. The latent space representations, with shape (1 x 256), were also stored during this process. Furthermore, they were paired with a validity label, indicating if the geometry kernel succeeded in decoding the sequence of commands into a B-rep. The final shape, after the labeling, had a shape of (1 x 257).

\begin{figure}[t!]
\begin{center}
\includegraphics[scale=.3]{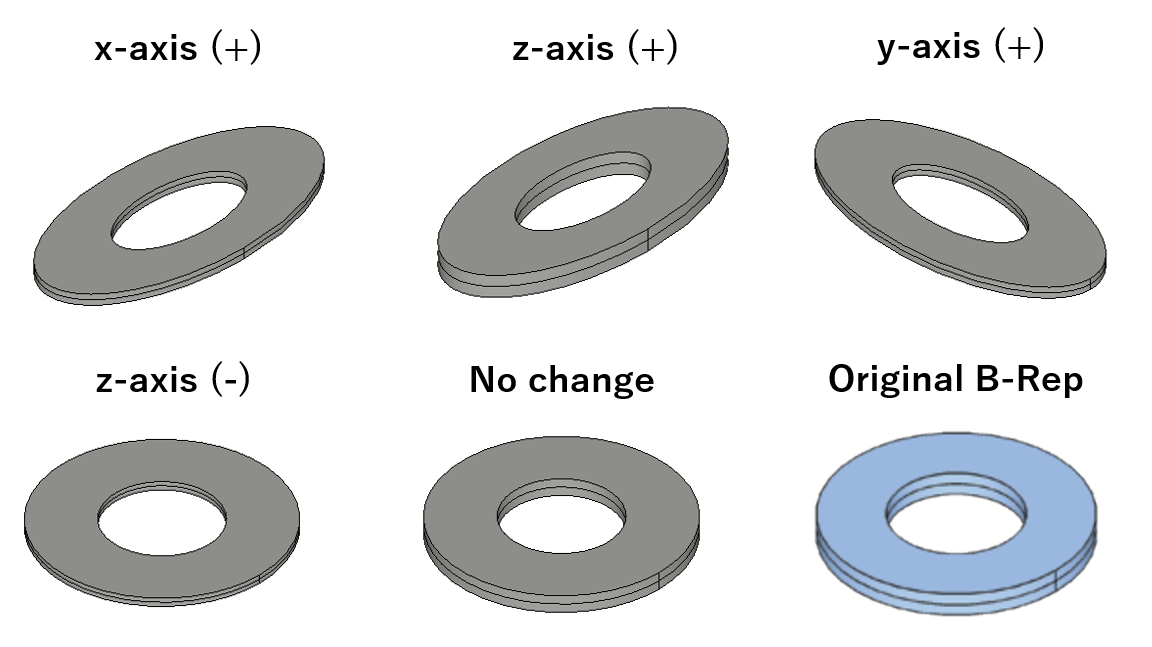}
\end{center}
\caption{Five-image translation per B-rep: x(+), y(+), z(+), z(-), no change.}
\label{fig:image_dataset} 
\end{figure}

Finally, the corresponding ground truth command sequences from the selected images were extracted. Using the GenCAD transformer encoder, they were converted into latent representations and saved as part of the data set. For every five images, a single command sequence was extracted, as they were generated by applying specific deformations to the same underlying B-rep. These deformations included translations along the x-axis (+), y-axis (+), z-axis (+), z-axis (-), and one unaltered version as illustrated in Fig.~\ref{fig:image_dataset}.

\subsection*{3.2. Model Components}
The Latent CAD Classifier and SSL Regressor were trained for use in the diffusion guidance and self-repair pipeline.


\subsection*{3.2.1. Latent CAD Classifier}
\begin{figure}[t!]
\begin{center}
\includegraphics[scale=.33]{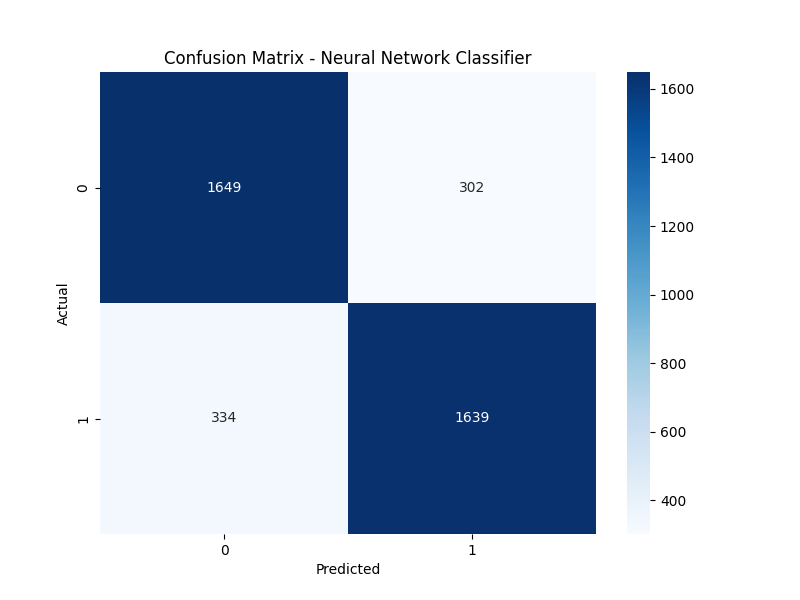}
\end{center}
\caption{Confusion matrix of the Latent CAD Classifier (1: valid, 0: invalid).}
\label{fig:confusion matrix} 
\end{figure}

The Latent CAD Classifier is a Multi-Layer Perceptron (MLP) trained on the GenCAD-Self-Repairing dataset, designed to classify an input latent vector as either valid or invalid.
To achieve class balance and mitigate overfitting, the valid latent vectors were undersampled. the Latent CAD Classifier is designed with an architecture consisting of three layers: 128, 64, and 1 neurons, respectively, with an input dimension of 256. Rectified Linear Unit (ReLU) activation functions were applied after the first two layers to introduce non-linearity, while a Sigmoid activation function was used in the output layer to produce probabilities for binary classification. The model was trained using an $80\%$ training and $20\%$ testing split, where the latent vectors served as inputs and their corresponding valid/invalid labels as outputs. The results of the classifier are shown in Table~\ref{tab:ASME4Classifier}, in addition to a confusion matrix in Fig.~\ref{fig:confusion matrix}. 

Notably, this model demonstrated the separability of the classes by analyzing the latent representations, which is fundamental for the approach taken by GenCAD-Self-Repairing.
\begin{table}[t!]
\caption{Performance of the Latent CAD Classifier.}
\begin{center}
\begin{tabular}{c c c c}
\label{tab:ASME4Classifier}
Class & Precision & Recall & F1-Score\\
\hline
1 = Valid & 0.83 & 0.85 & 0.84\\
0 = Invalid & 0.84 & 0.83 & 0.84\\
\hline
 &  & Accuracy & 0.84\\
\end{tabular}
\end{center}
\end{table}

\subsection*{3.2.2. Self-Supervised Learned Regressor} \label{subsub:SSLReg}
\begin{figure}[t!]
\begin{center}
\includegraphics[scale=0.265]{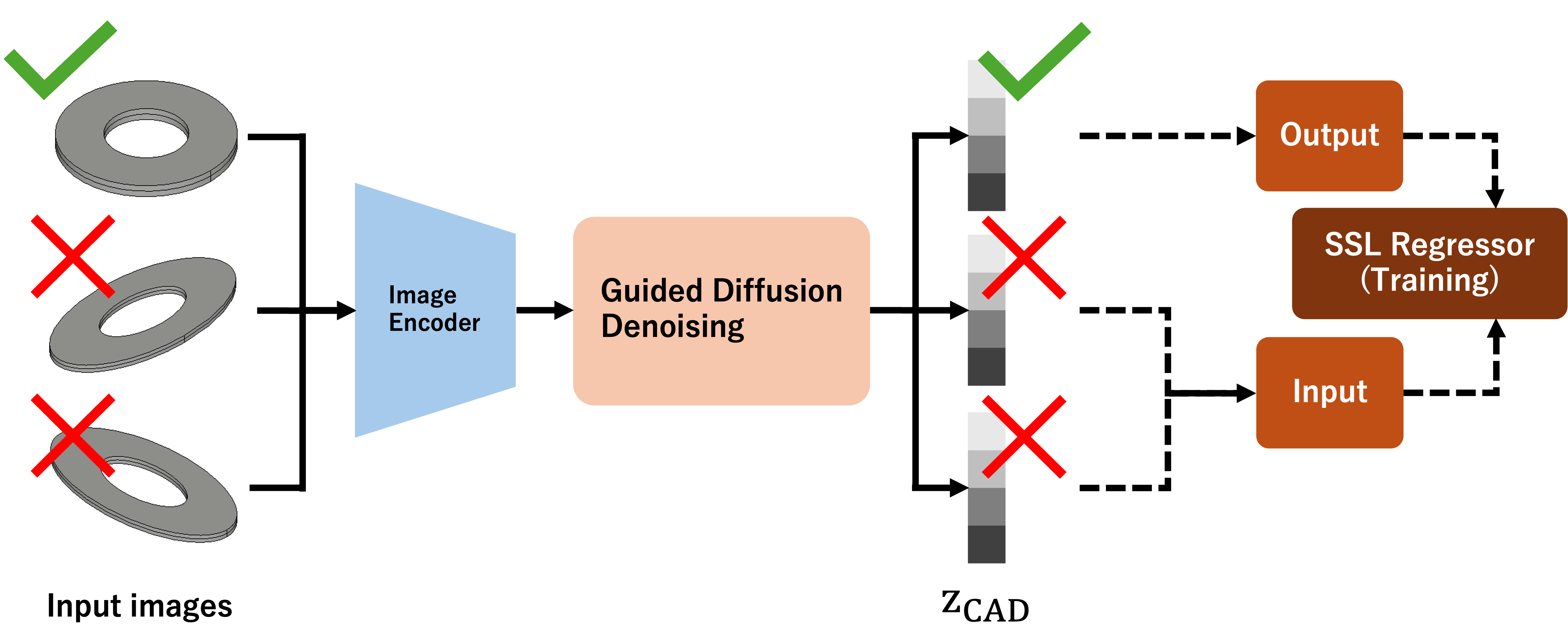}
\end{center}
\caption{Training flow of the SSL Regressor. It learned to map invalid latent representations to their corresponding valid latent representations, enabling self-repair in the latent space.}

\label{fig:SSLRegIO} 
\end{figure}

To train the SSL Regressor in a Self-Supervised  Learning manner, invalid latent representations of a given image are used as inputs, while the corresponding valid latent representations from the same image serve as outputs, as shown in Fig.~\ref{fig:SSLRegIO}. In essence, this is trying to predict what the valid representation should be given an invalid representation. 

It is important to note here that not all the images with invalid latent vectors were paired with at least one valid latent vector. After processing the data in this manner, there were a total of 9,029 pairs of invalid-valid latent vectors.  The SSL Regressor, along with all the other models, was implemented in Python using scikit-learn~\cite{scikit-learn}. With a multi-dimensional linear regression model, training $R^2 = 0.1275$ and $MSE = 0.0193$, and testing $R^2 = 0.0685$ and $MSE = 0.0206$ were obtained. 


This method, most notably, allowed for regression on the latent space without having to use the ground truth latent representations to define the desired outputs, in a fashion of Self-Supervised Learning. Furthermore, its use as a self-repairing mechanism did not alter the general structure of the original latent space, as discussed later.

\subsection*{3.3. Feasibility Enhancement Framework}
The proposed feasibility enhancement framework contains diffusion guidance and a self-repair pipeline by leveraging the trained Latent CAD Classifier and SSL Regressor.


\subsection*{3.3.1. Guided Diffusion Denoising}

Guided diffusion denoising is a key step in the proposed pipeline, leveraging both the Latent CAD Classifier and the SSL Regressor to guide the denoising process in latent space to pull invalid representations to valid regions (Fig.~\ref{fig:overview}). This diffusion guidance is applied to the diffusion prior (yellow prior in Fig.~\ref{fig:GencadArchitecture}), which transforms image embeddings encoded by the image encoder (blue encoder in Fig.~\ref{fig:GencadArchitecture}) into latent representations that can be decoded by the transformer decoder (gray decoder in Fig.\ref{fig:GencadArchitecture}). Note that these neural networks are the same as those used in \cite{gencad}.

This approach refines the denoising trajectory by incorporating gradient-based feedback from the Latent CAD Classifier and the SSL Regressor, ensuring that the generated samples are closer to the desired distribution.

The diffusion model computes the denoised latent vector $\mathbf{z}_{t-1}$ at each timestep $t$ through the following process:
\begin{align}
    \mathbf{\mu}_t &= \mathbb{E}[\mathbf{z}_{t-1} | \mathbf{z}_t, \mathbf{z}_0], \\
    \mathbf{z}_{t-1} &= \mathbf{\mu}_t + \sigma_t \mathbf{\epsilon},
\end{align}
where $\mathbf{\mu}_t$ is the posterior mean, $\sigma_t$ is the posterior variance, and $\mathbf{\epsilon} \sim \mathcal{N}(0, \mathbf{I})$ is Gaussian noise.

\paragraph*{Classifier Guidance:}
To ensure that the denoised sample aligns with the desired classification boundary, i.e., valid labeled regions, the Latent CAD classifier outputs the feasibility probability $P_{\mathrm{feasible}}(\mathbf{z}_t)$ at each timestep $t$.
We denote its infeasibility probability as $P_{\mathrm{inf}}(\mathbf{z}_t) = 1 - P_{\mathrm{feasible}}(\mathbf{z}_t)$, and use its gradient with respect to the current latent vector $\mathbf{z}_t$ to adjust the denoising direction. This guidance gradient is then subtracted from the posterior mean $\mathbf{\mu}_t$ to encourage feasible solutions:
\begin{align}
    \mathbf{\mu}_t' &= \mathbf{\mu}_t - s_{\mathrm{clf}} \nabla_{\mathbf{z}_t} P_{\mathrm{inf}}(\mathbf{z}_t)
\end{align}
where $s_{\mathrm{clf}} = 10$ is the scaling factor for the classifier guidance.


\paragraph*{Regressor Guidance:}
The SSL Regressor predicts a target value $\mathbf{y}_t$ that the latent vector $\mathbf{z}_t$ should approximate. To achieve this, a mean-squared error loss between the regressor output $\mathbf{y}_t$ and $\mathbf{z}_t$ is computed:
\begin{align}
    \mathcal{L}_\text{reg}(\mathbf{z}_t) &= \| \mathbf{y}_t - \mathbf{z}_t \|_2^2
\end{align}
The gradient of this loss with respect to $\mathbf{z}_t$ is scaled by the scaling factor $s_{\mathrm{reg}} = 10$ and used to further guide the denoising process based on the updated posterior mean $\mathbf{\mu}_t'$ guided by the classifier.
Specifically, building on $\mathbf{\mu}_t'$, the posterior mean $\mathbf{\mu}_t$ is further refined as:
\begin{align}
    \mathbf{\mu}_t^{\prime\prime} &= \mathbf{\mu}_t^\prime - s_{\mathrm{reg}} \nabla_{\mathbf{z}_t} \mathcal{L}_\text{reg}(\mathbf{z}_t).
\end{align}


\paragraph*{Final Update:}
The guided posterior mean $\mathbf{\mu}_t^{\prime\prime}$ is then used to compute the denoised sample:
\begin{align}
    \mathbf{z}_{t-1} &= \mathbf{\mu}_t^{\prime\prime} + \sigma_t \mathbf{\epsilon}.
\end{align}

This guided denoising process ensures that the diffusion model generates samples that align with the desired distribution. Specifically, it focuses on producing samples within the feasible region by satisfying the constraints imposed by the classifier and regressor, thereby improving the overall feasibility and reliability of the generated commands.

\subsection*{3.3.2. Self-Repair Pipeline}
After an image is processed with diffusion guidance and converted to a command sequence, the geometry kernel checks whether it can be converted into a B-rep. To verify the validity of a generated command sequence, we utilize Open CASCADE as the geometry kernel. The conversion attempt is performed by constructing a B-rep from the command sequence. If B-rep construction fails, the kernel deems the command sequence infeasible, and it is labeled as invalid. In that case, the latent vector is used as input for the SSL Regressor in an attempt to predict a feasible latent vector. The corrected output latent vector is finally decoded again, with the generated command sequence now having a better chance of corresponding to a valid geometry. This pipeline is shown in Fig.~\ref{fig:overview}.



\section{results and discussion} \label{sec:ResultsAndDiscussion}

The GenCAD validation set, consisting of 8,515 B-rep images, was used to evaluate GenCAD-Self-Repairing by comparing it against the baseline GenCAD model~\cite{gencad} and five variants of GenCAD-Self-Repairing, where different components were modified.

First, we describe the five GenCAD-Self-Repairing variants and their component modifications. Then, we present the evaluation metrics used for benchmarking these models.

\begin{figure}[t!]
\begin{center}
\includegraphics[scale=0.365]{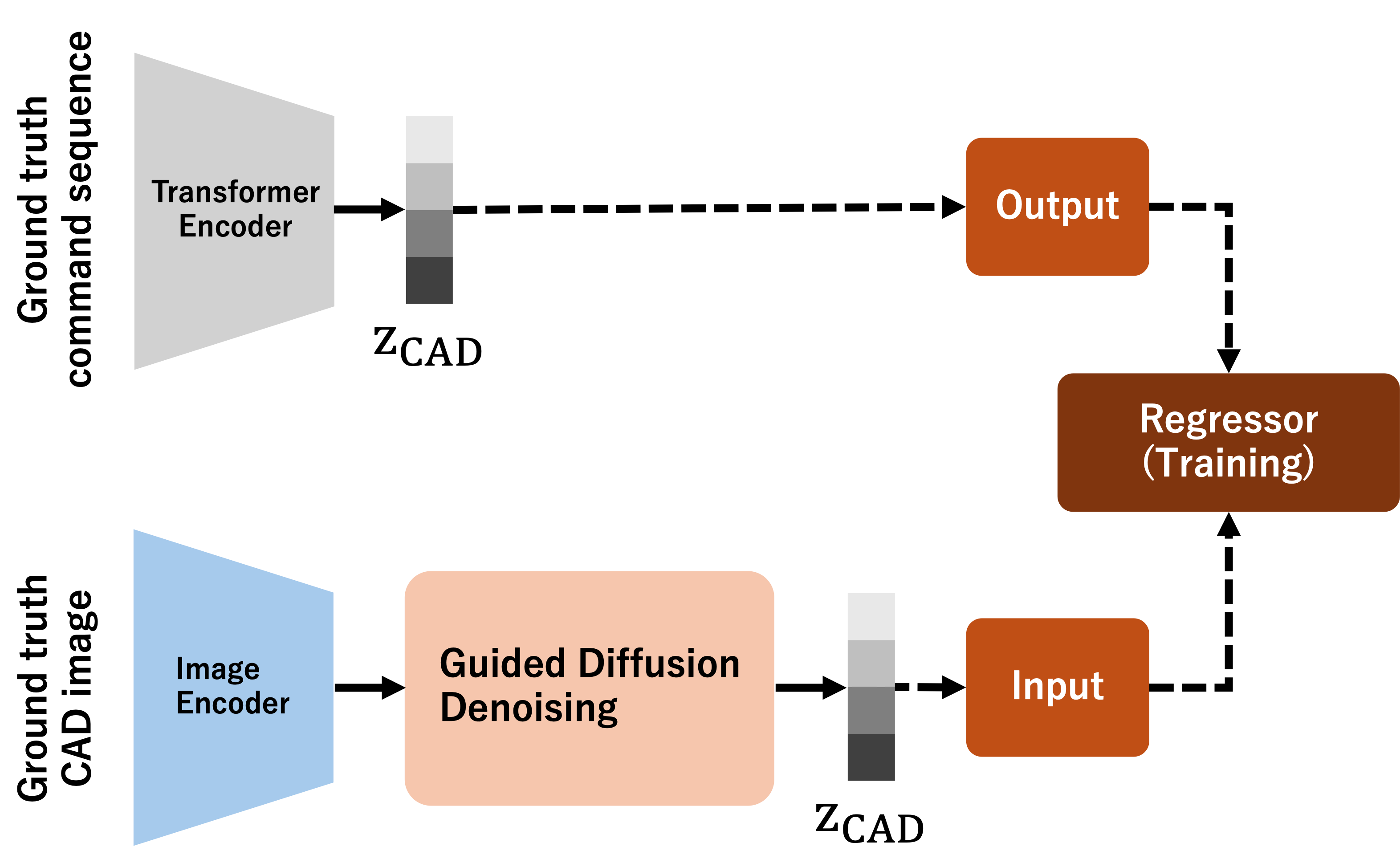}
\end{center}
\caption{Training flow of the Regressor in Var2. It learned to map generated latent vectors to their corresponding ground truth latent vectors.}

\label{fig:regressor2_train} 
\end{figure}


\subsection*{4.1. Variants of GenCAD-Self-Repairing}
We prepared five variants of GenCAD-Self-Repairing (Var1–Var5) to evaluate the effectiveness of each component and model selection: 
\begin{enumerate}
    \item Var1: Self-repair pipeline with the SSL Regressor.
    \item Var2: Self-repairing pipeline with the Regressor.
    \item Var3: Guided diffusion with the Latent CAD Classifier.
    \item Var4: Guided diffusion with the SSL Regressor.
    \item Var5: Guided diffusion with the Latent CAD Classifier and SSL Regressor.
\end{enumerate}

Here, GenCAD-Self-Repairing can be described as Guided diffusion with the Latent CAD Classifier and the SSL Regressor and Self-repair pipeline with the same SSL Regressor.

In Var2, we replaced the SSL Regressor described in Section~\ref{subsub:SSLReg} with a different regressor (Regressor).
The Regressor was trained on the GenCAD-Self-Repairing dataset using generated latent vectors as inputs and the ground truth latent vectors as outputs, as shown in Fig.~\ref{fig:regressor2_train}. The Regressor, in opposition to the SSL Regressor, is trying to modify the latent structure to achieve higher overall point cloud accuracy (MMD score) when compared to the underlying point clouds while simultaneously improving the feasibility rate.
Given that there was an available ground truth latent vector for every image, 133,617 pairs of vectors were used to fit the Regressor. The implementation was done with a multi-dimensional linear regression model. Training $R^2 = 0.6301$ and $MSE = 0.00414$, and testing $R^2 = 0.6268$ and $MSE = 0.00422$ were obtained. 


\subsection*{4.2. Evaluation Metrics}
The main goal was to achieve a higher feasibility rate. Hence, Eq.~\eqref{eq:first_matric} was the first evaluation metric utilized. The formal definition is given by: \begin{equation}
    F = \frac{V}{V + I}  
    \label{eq:first_matric}
\end{equation}
Where $F$ is the feasibility rate, $V$ is the total number of valid generated B-reps, and $I$ is the total number of invalid generated B-reps.

The second goal was to maintain the accuracy of the generated geometries. Several metrics can be used for this goal, such as Jensen–Shannon Divergence (JSD) or Maximum Mean Discrepancy (MMD). 

MMD is a kernel-based metric that measures the distance between two distributions, $P(X)$ and $Q(Y)$, by embedding them in the Reproducing Kernel Hilbert Space (RKHS) and calculating the distance between their means ~\cite{mmd}. The formula is given as:   
\begin{equation}
\mathrm{MMD}(P, Q) = \left\| \mu_X - \mu_Y \right\|_{\mathcal{H}}
\end{equation}

JSD also measures the statistical similarity between distributions. Nevertheless, JSD discretizes the data by computing a normalized histogram of the points in the point clouds, losing information in the process of binning ~\cite{jsd}. MMD, on the other hand, uses all the datapoints to compare the means between the distributions of the point clouds, being better suited for complex distributions but also more computationally expensive. 

Given the unknown underlying distributions, MMD was chosen as the accuracy metric. The kernel function utilized was the Gaussian Radial Basis Function (RBF): 
\begin{equation}
    k(x,y) = e^{-\frac{||x-y||^2}{2\sigma^2}}  
\end{equation}

This was because we assumed that close points are more highly correlated than points that are further apart. Furthermore, since the means for the point cloud distributions are unknown, the MMD metric is calculated empirically as:
{\begin{equation}
\small
\text{MMD} = \sqrt{\frac{1}{m^2} \sum_{i,j=1}^{m} k(x_i, x_j) 
+ \frac{1}{n^2} \sum_{i,j=1}^{n} k(y_i, y_j) 
- \frac{2}{mn} \sum_{i=1}^{m} \sum_{j=1}^{n} k(x_i, y_j)}
\end{equation}
}


\subsection*{4.2. Benchmark Evaluation}
The Table \ref{results} shows the feasibility rate  $F$ 
 and the average MMD Score for all the described methods over the validation set of images. The best results are shown in \textbf{bold letters}. (↓) indicates that lower is better, and (↑) indicates that higher is better.

\begin{table}[b]
\caption{Benchmark results comparing GenCAD-Self-Repairing with variants and the baseline.}
\resizebox{0.995\linewidth}{!}{
\centering

\begin{tabular}{p{3.8cm}|c|c}  
\hline
Method & Feasibility (↑) & Mean MMD (↓) \\
\hline
GenCAD (baseline)~\cite{gencad} & 0.931 & \textbf{0.180} \\
Var1 & 0.961 & 0.183 \\
Var2 & 0.931 & 0.181 \\
Var3 & 0.936 & 0.197 \\
Var4 & 0.936 & 0.198 \\
Var5 & 0.954 & 0.199 \\
\makebox[0pt][l]{\textbf{GenCAD-Self-Repairing}} & \textbf{0.970} & 0.200 \\
\hline
\end{tabular}
}

\label{results}
\end{table}

\begin{figure}[t!]
\begin{center}
\includegraphics[scale=0.46]{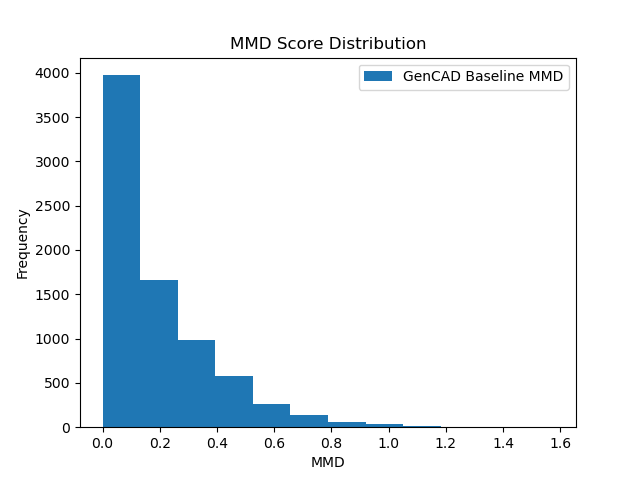}
\end{center}
\caption{histogram of GENCAD BASELINE MMD SCORES.}
\label{fig:baseline_mmd} 
\vspace{-4mm}
\end{figure}

\begin{figure}[t!]
\begin{center}
\includegraphics[scale=0.46]{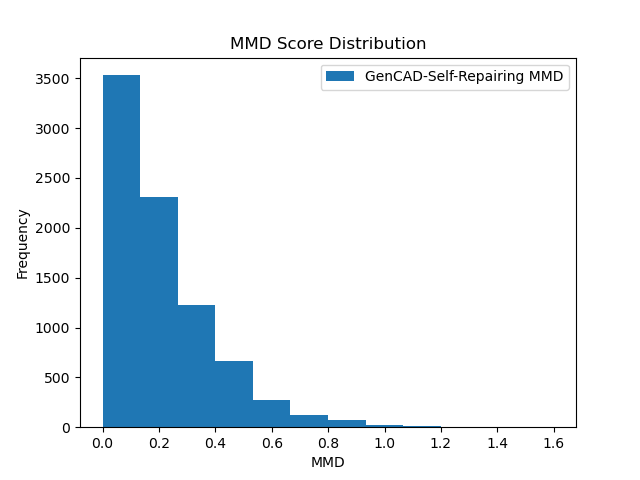}
\end{center}
\caption{histogram of GENCAD-SELF-REPAIRING MMD SCORES.}
\label{fig:self_repair_mmd} 
\vspace{-4mm}
\end{figure}

GenCAD (baseline)~\cite{gencad} was able to convert 7,707 B-reps from 8,515 images, whereas GenCAD-Self-Repairing converted 8,239 B-reps from 8,515 images. This means that GenCAD-Self-Repairing successfully fixed 65.84\% ($532/808$) of the baseline infeasible images, significantly improving feasibility. 


However, there was an 11.11\% increase in MMD ($0.2/0.18$), meaning that the overall accuracy slightly decreased compared to the baseline. Examining the distribution of MMD scores, as shown in Fig.~\ref{fig:baseline_mmd} and Fig.~\ref{fig:self_repair_mmd}, the histograms illustrate that while the overall distribution shape remained similar, there was a notable decrease in the leftmost bar (lower MMD values) and an increase in the second-left bar. This suggests that GenCAD-Self-Repairing exhibited a slight reduction in accuracy for simpler shapes, which were more accurately reconstructed by the baseline.

\begin{figure}[t!]
\begin{center}
\includegraphics[scale=0.48]{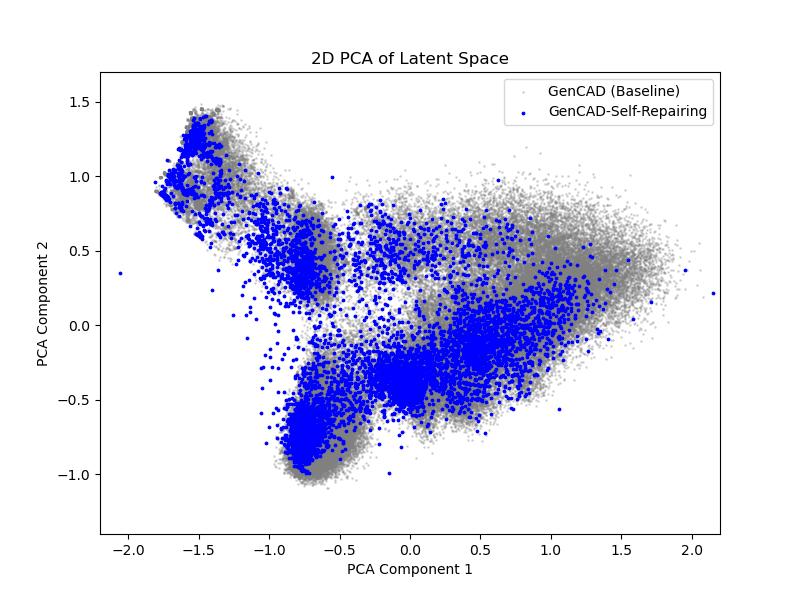}
\end{center}
\caption{2D PCA OF THE LATENT SPACE both the GenCAD baseline~\cite{gencad} (gray) and GenCAD-Self-Repairing (blue).}
\label{pca} 
\end{figure}

Finally, we analyzed the 2D Principal Component Analysis (PCA)~\cite{1261097} of the latent space to evaluate the impact of GenCAD-Self-Repairing on the underlying distribution. As shown in Fig.~\ref{pca}, the latent space distribution of GenCAD-Self-Repairing was closely aligned with that of the baseline GenCAD model~\cite{gencad}. This indicates that the proposed feasibility enhancement framework effectively corrected infeasible latent representations without significantly altering the overall latent structure.
By preserving a similar latent space, GenCAD-Self-Repairing improved feasibility while maintaining the generative behavior of the original model. This suggests that the method successfully guided infeasible samples toward feasible regions without disrupting the design space, enhancing CAD generation success rates without introducing unintended distribution shifts.

\section{CONCLUSION and FUTURE WORK} \label{sec:ConculsionAndFuturework}

The advancements in generative AI are gradually reshaping the CAD modeling process, with an increasing number of AI-driven models focusing on CAD generation from images, text embeddings, and command sequences. However, most existing models have fixed accuracy and feasibility rates, limiting their adaptability and applicability to different tasks.

In this work, we introduced GenCAD-Self-Repairing, a method designed to increase the feasibility of CAD generation by guiding the diffusion denoising process, and correcting infeasible command sequences with the self-repair pipeline. Our approach successfully converted two-thirds of infeasible designs into feasible ones, significantly improving the feasibility rate compared to the baseline GenCAD model~\cite{gencad}.

However, one limitation of GenCAD-Self-Repairing is the slight decrease in accuracy, as indicated by an 11.11\% increase in MMD compared to the baseline. The histogram analysis suggests that the decrease in accuracy primarily affects simpler shapes, which the baseline model handled with higher precision. This tradeoff between feasibility and accuracy highlights an area for further improvement. 

Future work will focus on exploring more advanced models, such as transformers that incorporate time-series information, to refine latent space corrections and mitigate accuracy degradation. Additionally, integrating adaptive guidance mechanisms within the diffusion process may further enhance feasibility while preserving accuracy.

\section*{Acknowledgments}
We would like to thank Professor Faez Ahmed, and Ph.D. Candidate
Lyle Regenwetter for their mentorship and advice. Their support was truly outstanding.


\nocite{*}

\bibliographystyle{asmeconf}  
\bibliography{asmeconf-sample}


\end{document}